\documentclass[10pt,twocolumn,letterpaper]{article}

\usepackage[pagenumbers]{iccv} 

\usepackage{pifont}
\usepackage{algorithm}
\usepackage{algpseudocode}
\usepackage{appendix}

\definecolor{iccvblue}{rgb}{0.21,0.49,0.74}
\usepackage[pagebackref,breaklinks,colorlinks,allcolors=iccvblue]{hyperref}

\def\paperID{13273} 
\def\confName{ICCV}
\def\confYear{2025}

\title{Close-up-GS: Enhancing Close-Up View Synthesis in 3D Gaussian Splatting with Progressive Self-Training}
\author{Jiatong Xia\quad \quad~~ Lingqiao Liu\thanks{}\\
Australian Institute for Machine Learning, The University of Adelaide\\
{\tt\small \{jiatong.xia,~lingqiao.liu\}@adelaide.edu.au}
}

\begin{document}

\twocolumn[{%
\centering
\renewcommand\twocolumn[1][]{#1}%
\maketitle
    \captionsetup{type=figure}
    \vspace{-3mm}
    \includegraphics[width=0.94\textwidth]{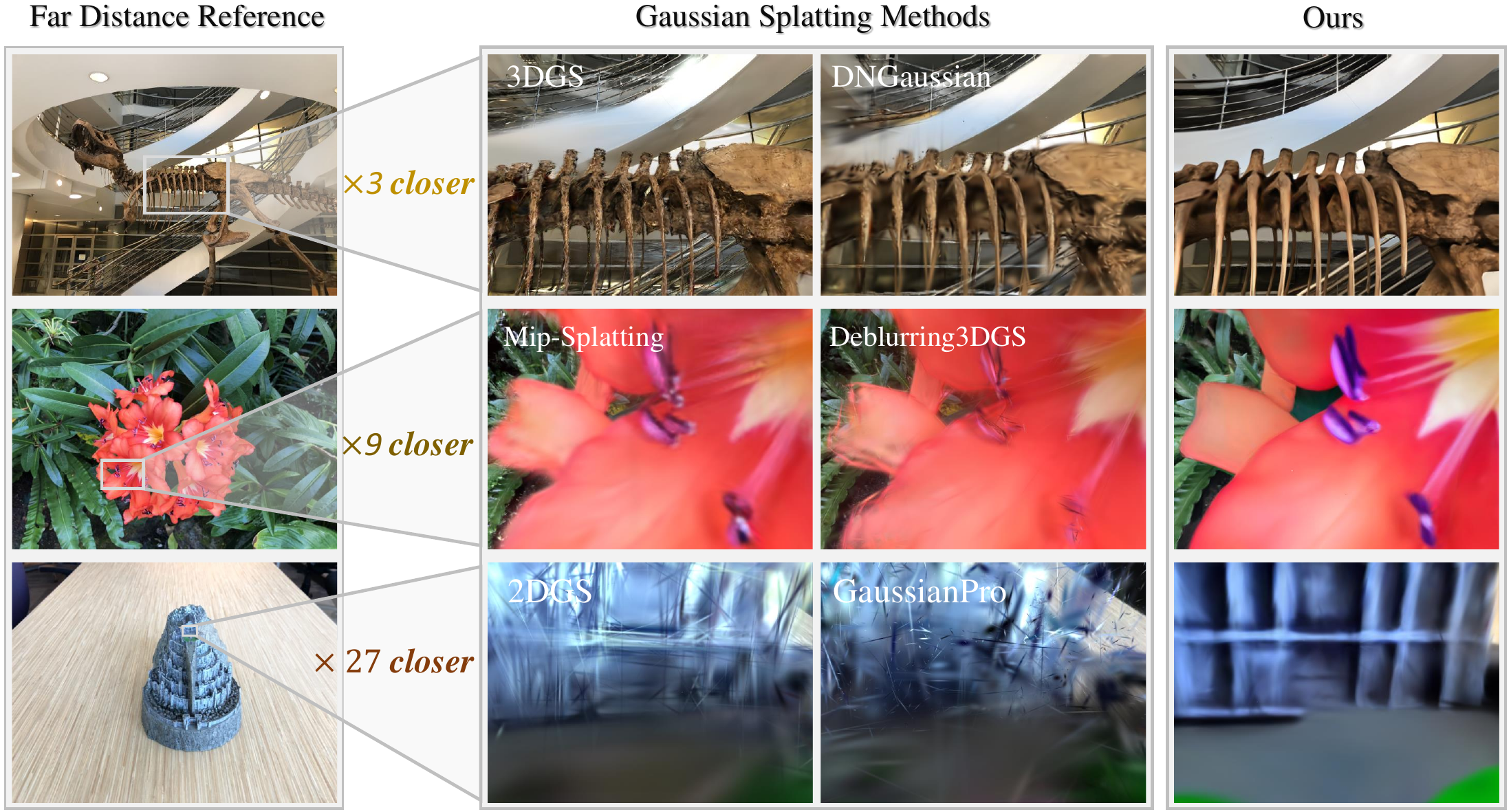} 
    \vspace{-1mm}
    \hfill\caption{\textbf{Close-up view synthesis} 
    remains a challenge in Gaussian Splatting, often lacking effective solutions. In this paper, we aim to address the close-up views issue by proposing a novel method Close-up-GS. Our method demonstrates impressive rendering quality across different close-up scales, even at a 27$\times$ close-up distance, achieving extreme close-up with remarkable details.}
    \label{fig:teaser}
    \hfill 
    \vspace{3mm}
}]

\begin{abstract}
3D Gaussian Splatting (3DGS) has demonstrated impressive performance in synthesizing novel views after training on a given set of viewpoints. However, its rendering quality deteriorates when the synthesized view deviates significantly from the training views. This decline occurs due to (1) the model's difficulty in generalizing to out-of-distribution scenarios and (2) challenges in interpolating fine details caused by substantial resolution changes and occlusions. A notable case of this limitation is close-up view generation—producing views that are significantly closer to the object than those in the training set. To tackle this issue, we propose a novel approach for close-up view generation based by progressively training the 3DGS model with self-generated data. Our solution is based on three key ideas. First, we leverage the See3D model, a recently introduced 3D-aware generative model, to enhance the details of rendered views. Second, we propose a strategy to progressively expand the ``trust regions'' of the 3DGS model and update a set of reference views for See3D. Finally, we introduce a fine-tuning strategy to carefully update the 3DGS model with training data generated from the above schemes. We further define metrics for close-up views evaluation to facilitate better research on this problem. By conducting evaluations on specifically selected scenarios for close-up views, our proposed approach demonstrates a clear advantage over competitive solutions.
\end{abstract}    
\section{Introduction}
\label{sec:intro}
Advances in radiance field methods have enabled remarkable progress in synthesizing novel views of a 3D scene from input training views. Among these approaches, recent work on 3D Gaussian Splatting (3DGS)~\cite{kerbl3Dgaussians} has drawn significant attention for its ability to produce high-quality renderings in real time. By modeling a scene with an explicit set of 3D Gaussians, and optimizing a radiance field over these primitives, 3DGS reduces both training and inference costs compared to implicit representations (e.g., NeRF~\cite{mildenhall2020nerf}). Through an efficient rasterization rendering pipeline, 3DGS has demonstrated impressive fidelity and speed for novel view synthesis.

Despite its strengths, 3DGS struggles when required to predict appearances under out-of-distribution conditions. In particular, close-up view synthesis—where the camera moves significantly closer to an object than any of the training viewpoints—leads to notable degradations. Two issues predominantly underlie this limitation. First, a large fraction of the rays in these novel views diverge substantially from what the model has seen during training, leading to artifacts in the color rendering. Second, any fine details that were interpolated or occluded in the training views are difficult to restore from the 3DGS alone. Currently, there is still a gap in research to address all these issues well and provide an effective solution to the close-up views problem.

To address this challenge, we propose a progressive self-training framework for close-up view synthesis, guided by three core ideas that build upon the strengths of 3D Gaussian Splatting (3DGS) while overcoming its limitations in out-of-distribution scenarios. First, we leverage See3D, a newly developed 3D-aware generative model that can generate novel views with the guidance of existing views, to refine the missing details in newly rendered close-up views. See3D serves as a data-driven prior, injecting intricate details that would otherwise remain absent when the camera moves significantly closer than the training distribution allows\footnote{Note using the generative model inevitably introduces content that might not always match the unobserved ground-truth. In this work, our aim is to create a close-up view that balances the fidelity and visual quality. }. Second, we introduce a progressive update scheme to expand the 3DGS “trust region,” which systematically identify and select intermediate novel views between the far-distance training set and the final close-up vantage points. This approach mirrors the idea of pseudo-labeling in semi-supervised learning: some refined (and verified) output becomes a new training sample that helps 3DGS adapt to increasingly difficult viewpoints, effectively bridging the gap from far to near distances. Finally, we design a specialized fine-tuning procedure for 3DGS, ensuring consistent geometry and color mappings between real and self-generated images. By introducing new metrics focused on evaluating close-up render quality, we offer a thorough assessment of this method’s practical impact. Our experiments on challenging benchmarks confirm that this progressive self-training paradigm substantially outperforms existing solutions, extending the utility of Gaussian Splatting to scenarios that demand precise, high-quality results at extreme close-ups.

\section{Related Work}
\label{sec:related}

\paragraph{Radiance Fields.}
The development of radiance fields has significantly advanced the learning of 3D representations from 2D images. This type of methods were first introduced by Neural Radiance Fields~\cite{mildenhall2020nerf} (NeRF), which implicitly represent a scene using multi-layer perceptions (MLPs) and applying volume rendering for pixel rendering. Since then, NeRF-related techniques have been extensively adopted in various applications~\cite{kerr2023lerf,zhang2021ners,li2023uhdnerf}. 
However, NeRF has several limitations, primarily due to its implicit representation, which relies on neural networks for ray rendering, leading to high computational costs and inefficiencies in both training and inference. 
Several advancements~\cite{mueller2022instant, yu2021plenoctrees,chen2022tensorf} have been made, enabling a much faster process.
Compared to NeRF, 3D Gaussian Splatting (3DGS)~\cite{kerbl3Dgaussians} employs an explicit representation to model radiance fields, significantly reduces training time while enabling high-quality and real-time rendering.
With the widespread application of 3DGS, further improvements~\cite{cheng2024gaussianpro,lee2024deblurring,chen2024mvsplat,jiang2024gaussianshader} have been proposed to enhance its ability in different scenarios. Yuan et al.~\cite{Huang2DGS2024} introduced 2D Gaussian Splatting (2DGS), which provides accurate and view-consistent geometry, and enhance the quality of the reconstructions.
Yu et al.~\cite{Yu2024MipSplatting} addressed aliasing issues by designing sampling filters based on sampling principles. 
These Gaussian Splatting related method advancements significantly enhanced radiance fields, enabling them to achieve remarkable performance in faithfully reconstructing real-world scenes.
\vspace{-3pt}

\paragraph{Unconventional Condition Radiance Fields.}
As Gaussian splatting and other radiance field methods are applied to real-world scenarios, various challenges have emerged under unconventional conditions, leading to targeted improvements. For example, when the number of training views is insufficient, the rendering results will suffer from low quality.
Several methods~\cite{deng2022depth, roessle2022dense, kim2022infonerf, niemeyer2022regnerf} first addressed this issue in NeRF, and more recently, methods~\cite{xiong2023sparsegs,liu20243dgs-enhencer} like DNGaussian~\cite{li2024dngaussian} tackle this problem in 3DGS.
Another situation different from limited views is single-view reconstruction; many works~\cite{szymanowicz2024splatter_image,zou2023triplane,xu2024agg} have been proposed to address this issue.
As the study of radiance fields deepens, it has also been discovered that even with sufficient training samples, issues arise when rendering from viewpoints that deviate from the training views, this is known as an out-of-distribution (OOD) issue. SplatFormer~\cite{chen2024splatformer} introduced this issue in general 3DGS. They use low-elevation views as training views and enhance 3DGS in the high-elevation views rendering, yet it does not specifically address the close-up view problem.
Xia et al.~\cite{xia2025pseudo} first highlight the OOD problem when moving the camera closer and proposed a method based on geometric consistency to generate reliable sparse pixels for those views. However, their approach does not address issues such as occluded regions or interpolated details at close distances. 
In general, there is still a lack of effective solutions to address rendering issues in close-up views.
\vspace{-6pt}

\paragraph{Diffusion Priors.}
Applying diffusion models~\cite{ho2020ddpm,song2020ddim,zhang2023adding,ho2022video} to generate reliable images or enhance existing images~\cite{meng2022sdedit,yu2024supir} has become a well-established technology.
Due to the scalability of diffusion models, their effectiveness has been evolving at a fast speed.
Using the priors of diffusion models to obtain 3D representations such as radiance fields, has become a key area of research.
Many recent works~\cite{poole2022dreamfusion,tang2023dreamgaussian,liang2024luciddreamer} apply score distillation sampling (SDS) to distill 2D diffusion priors into 3D representations, however, they currently limited to object-level reconstruction.
A recently proposed See3D~\cite{Ma2025See3D} applies a reference-guided, warping-based generation framework. It utilizes known images as references and warped results of known images under novel views as input to restore the novel view results, demonstrating strong performance in generating novel views for complex scenes.
Thus, using See3D's priors to enhance radiance fields holds significant potential, while it has many challenges.
3D representations such as 3DGS impose strict constraints on geometric consistency, and subtle inconsistencies can impact the rendering quality.
See3D can maintain some level of consistency, but it faces challenges in achieving a fully consistent reconstruction across all views, particularly in close-up scenarios.
In this paper, we aim to propose a robust method that leverages See3D's priors to address the challenges of close-up views in 3DGS.
\section{Preliminary: Gaussian Splatting}
\label{sec:preliminary}

\label{pre_3dgs}
Gaussian splatting \cite{kerbl3Dgaussians} represents a 3D scene using anisotropic 3D Gaussians and enable image rendering through a differentiable tile rasterization process.
Each 3D Gaussian primitive $G_k(p)$ at a center point $p_k=\left \{ x_k, y_k, z_k \right \}$ is described together with a covariance matrix $\Sigma_k$, the function of the $k$-th primitive can be formulate as:
\begin{equation}
\begin{split}
 G_k(p) = exp(-\frac{1}{2}(p-p_k)^T\Sigma_k^{-1}(p-p_k)),
    \label{eq:gaussian}
\end{split}
\end{equation}
To render an image, 3D Gaussians are transformed into 2D camera coordinates using the world-to-camera transform matrix $W$ and the Jacobian $J$ of the affine approximation of the projective transformation~\cite{zwicker2001surface} as follows:
\begin{equation}
    \Sigma' = J W \Sigma W^{T} J^{T}.
    \label{eq:cov2d}
\end{equation}
A point-based rendering is utilized to compute the color $C$ of each pixel, by blending $\mathcal{N}$ depth-ordered Gaussians overlapping the pixel, with each Gaussian has its opacity $\sigma$ and color $c$ which is determined using the SH coefficient with the viewing direction. Formulate as:
\vspace{-2pt}
\begin{equation}
    C = \sum_{i \in \mathcal{N}} c_i \alpha_i \prod_{j=1}^{i-1} (1 - \alpha_j), \\
    \label{eq:blending}
\end{equation}
where $\alpha_i = \sigma_iG_i(p)$. The optimization process minimizes the difference between the rendered color and the ground truth images, using the following loss function:
\vspace{-1pt}
\begin{equation}
    \mathcal{L} =(1-\lambda )\mathcal{L}_{1} + \lambda \mathcal{L}_{SSIM}
\end{equation}
where ${\mathcal{L}}_{1}$ and $\mathcal{\mathcal{L}}_{SSIM}$ represent the image ${\mathcal{L}}_{1}$ loss and the similarity loss, respectively.

\vspace{-1pt}
\section{Methods}
\label{sec:method}

\subsection{Close-up Views Rendering Problems}

Our work focuses on the problem of generating novel close-up views around an area-of-interest given a set of far-distance training views.
Formally, we assume an initial 3DGS model has already been trained from a collection of known views $\mathcal{V}_{s} = \{V_i\},i=1\cdots n$. The center of object-of-interest is given as $p_{target}$\footnote{For the simplicity, we assume a single center for the object-of-interest. Our approaches can be straightforwardly generalized to the case with multiple centers if the object of interest involves multiple objects.}. The aim of our approach is to update the 3DGS model to render high-quality views around $p_{target}$.

This task is a non-trivial task due to two reasons: 
(1) The rays that can be accurately rendered by Gaussian Splatting are within the regions that can be interpolated from the rays in training views.
Close-up views inevitably include rays that deviate from the known regions, which makes it difficult for 3DGS to render accurately. A similar issue occurs when rendering rays pass through areas that training rays are occluded, the missing information in the occluded regions can cause artifacts in 3DGS rendering. (2) In addition, the finer details in the close-up region are interpolated from far-distance views, the limited resolution of these distant images often results in lower-quality close-up details.

\subsection{Enhancement with Generative Models}
To address the above limitation, this paper leverages a generative model to infer missing details using its rich prior knowledge. Specifically, we adopt the recently proposed See3D \cite{Ma2025See3D}, defined as $\mathcal{M}_{\theta}$, which utilizes reference views to guide novel view generation. It generates a novel view based on several inputs: (1) multiple reference views, (2) an initial image input of the novel view, which can be obtained by wrapping from reference views, and (3) a mask highlighting which pixels are unknown. Unlike standard image generative models, See3D generates with the guidance of other views and thus can better preserve the consistency across view. Moreover, its mask input allows user to define the known pixel and unknown pixel, making it ideal for the close-up rending problem as we wish the generation not deviating too much from the content can be rendered from existing 3DGS model.
Formally, See3D creates the novel view image via:
\begin{align*}
    I_n = \mathcal{M}_{\theta}(I'_n, M'_n, \{I^{ref}_k\})
\end{align*}

In our work, we propose a scheme to seamlessly integrate See3D to mitigate information loss in out-of-distribution view rendering. Given a novel view, we first render it using a pretrained 3DGS model and then identify reliable regions using the method from Xia et al.~\cite{xia2025pseudo}. This produces a mask highlighting pixels that are geometrically consistent with the training views, which serves as input for the See3D model. The model then refines the novel view, and its output can undergo further enhancement, such as super-resolution~\cite{yu2024supir}, for improved interpolated close-up details and quality.
Once the refined image is generated, it can be used to update the Gaussian Splatting by treating it as a new training view. 

Although conceptually simple, this approach faces some challenges when generating close-up views. 
To achieve high-quality generation, it is essential that the reference views provide sufficient coverage of the novel view, allowing the See3D model to fully utilize their guidance. Thus, the See3D model should be used to generate views that are not too far from the reference views.
This implies that the 3DGS model update process should be performed incrementally in multiple steps, and each step requires a carefully designed scheme to select the reference views and the views to be updated, ensuring both high generation quality and efficiency.
Additionally, the reference views should be continuously updated, and the update process of the 3DGS model must be specially designed to maintain consistency across views.

\subsection{Progressive model update scheme}
\begin{figure*}[!t]
\centering
\includegraphics[width=0.99\linewidth]{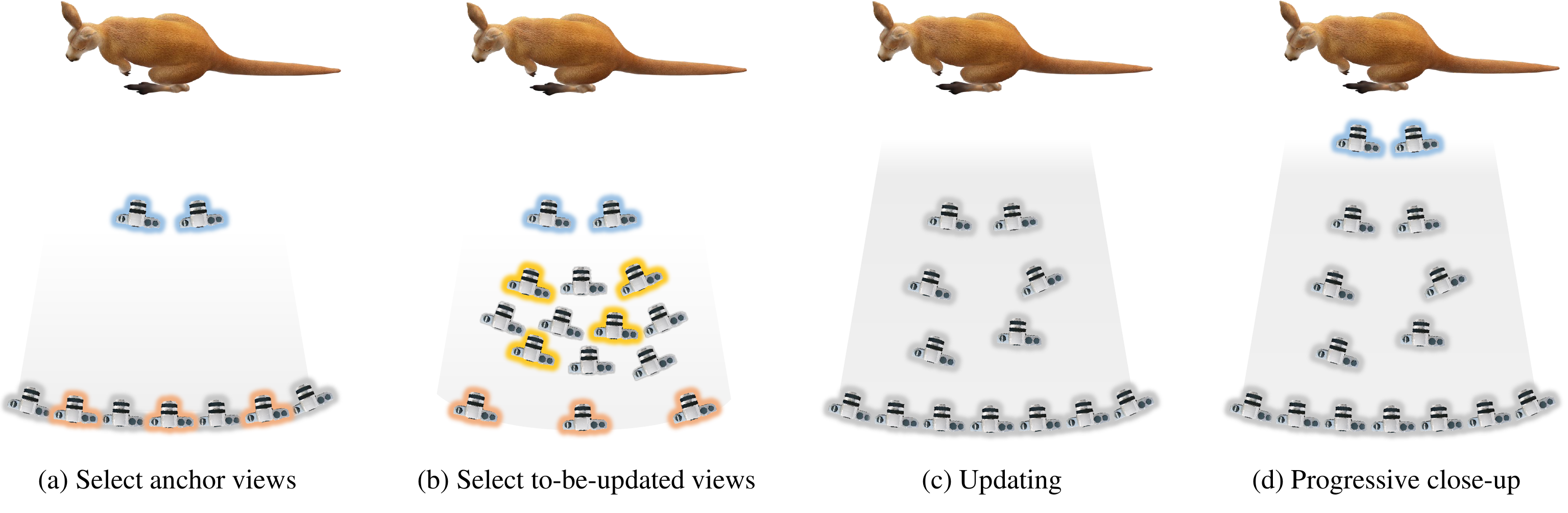} 
\vspace{-6pt}
\caption{Progressive update. (a) Setting the frontier views (blue camera) and select the anchor views (orange camera) from know views (gray camera) based on our views selection algorithm. The anchor views will also serve as the reference views of See3D. (b) Select the to-be-updated views (yellow camera) from a various of random views between frontier views and anchor views, also using the selection algorithm. (c) Render and refine the to-be-updated views and frontier views, then apply our fine-tuning procedure to update 3DGS therefore extend its ``trust region'' (gray region). The to-be-update views and frontier views are then added to the known views. (d) Begin the next round by setting new frontier views.
}
\vspace{-4pt}
\label{fig:progressive}
\end{figure*}

In this paper, we propose a scheme to progressively updating the 3DGS model by gradually expanding the ``trust region'' of the 3DGS, ie, the region in which the 3DGS model can generate high-quality views. This process involves multiple rounds of expansion, which we illustrate in Fig.~\ref{fig:progressive}. In each expansion, the scheme involves four steps: (1) select the anchor views $\mathcal{R}_s$ of the current known views. (2) Select the to-be-updated views $\mathcal{V}_u$ outside of of the current trust region. (3) render and refine the views to be updated using anchor views as references. (4) update the 3DGS model by fine-tuning it on the to-be-updated views and then append those views to the known views. The whole process is akin to the pseudo-labeling process in semi-supervised learning.

In our implementation, we separate the whole process into multiple expansion stages, based on the distance to the center of the object-of-interest $p_{target}$. The frontier of the trust region will get closer to $p_{target}$ in each round. More implementation details can be found in section \ref{sec:imple}.
In the following sections, we elaborate on the algorithm details for each step in the process.

\subsubsection{Select anchor views}
In each expansion round, our method selects $k$ anchor views from $n$ known views ($ k < n$). Those anchor views will be used later in the other steps, such as reference views for See3D, using $k$ anchors rather than all known views helps reduce computational cost. Intuitively, the selection of the anchor views should follow two principals: (1) the view should cover the object of interest as much as possible. (2) The redundancy of view should be minimized. 

To evaluate how well a view covers the object-of-interest, our method identifies a set of frontier views through the following process. Initially, we select an anchor view—either discovered in the previous round or randomly chosen in the first round—and draw a line segment connecting it to $p_{target}$ with a length of $d$. A virtual camera is then positioned along this line at a distance of $\alpha_t\cdot d$ from $p_{target}$, where $0 < \alpha_t < 1$ is a predefined constant for each round. The camera's orientation is subsequently adjusted to ensure that $p_{target}$ is projected at the center of the image. With $k$ anchor views, this process will produce $k$ frontier views $\{v^f_j\},j=1\cdots k$. These frontier views represent the closest views achievable within the distance constraints for camera placement in the current round. 

Using these frontier views, we can quantify how well a known view covers the target region. This is determined by calculating the number of pixels visible in both the frontier view and the known view\footnote{This can be achieved by warping the frontier view to align with the known view. In practice, higher weights are assigned to pixels near the center of the frontier view, as they are more relevant to the object-of-interest.}. Averaging the results from $k$ frontier views, we obtain a coverage score $w^i$ for the $i$-th known view. In a similar way, we can quantify the redundancy between two views by counting how many pixels are visible in both views. This produces a pairwise view similarity score $e_{i,j}$. Representing $\{w^i\}$ in a vector form as $\mathbf{w} \in \mathbb{R}^n$ and $\{e_{i,j}\}$ in a matrix form as $\mathbf{E} \in \mathbb{R}^{n \times n}$, anchor view selection can be formulated as an optimization problem:
\vspace{-6pt}
\begin{align}\label{eq:opt}
    \max_{\mathbf{s}} & ~\mathbf{s}^\top\mathbf{w} - \mathbf{s}^\top\mathbf{E}\mathbf{s} \nonumber \\
    & s.t. ~~ \mathbf{s}^\top\mathbf{1} = k
\end{align}
where $\mathbf{s} \in \mathbb{R}^n $ is a binary vector, with ``1'' element indicates select and ``0'' not select. This is a combinatorial optimization problem and exists greedy approximation solution. We leverage the solution which is detailed in the supplementary to calculate the optimal selection as $k$ anchor views.

\vspace{-1pt}

\subsubsection{Select to-be-updated views}
In a given expansion round, anchor views serve as a summary of the known views, while the frontier views represent the targets we aim to reach. Our progressive model update scheme selects a set of to-be-updated views located between the known views and the frontier views for rendering, refinement, and updating of the 3DGS model.
The selection of to-be-updated views begins by randomly sampling $M$ views whose distance to $p_{target}$ is smaller than that of the known views but larger than that of the frontier views. We then apply a similar procedure as used for selecting anchor views to obtain the final set of to-be-updated views. This process is illustrated in Fig. \ref{fig:progressive} (b).

Specifically, we still solve a similar optimization problem as in Eq. \ref{eq:opt} to select views and use the same way to calculate the similarity score. The only difference is when we calculate $w^j$, we also consider both the view overlap with the frontier view and with the anchor view. This is achieved by taking the average of the pixel overlap count calculated from both frontier views and with the anchor views. This approach ensures that the to-be-updated view serves as a ``bridge'' between the known view and the frontier view, preventing the See3D-generated view from deviating too far from the reference view—i.e., the anchor views in our framework. Additionally, we apply a discount factor, as detailed in the supplementary materials, to mitigate the influence of distance in the calculation.

\vspace{-1pt}

\subsubsection{Render and refinement}
Once the to-be-updated views are selected, we could run 3DGS model and See3D model to render and refine the views, where the $k$ anchor views will be used as references for See3D here. We also employ an image enhancement model \cite{yu2024supir} to further enhance the details and quality of the refined to-be-updated view images. 
Those view images are then used for updating the 3DGS model. These identified updated views are also be appended to the known views for the next round calculation.
\vspace{-1pt}

\subsubsection{Gaussian Splatting Fine-tuning}
The most straightforward approach to updating the 3DGS model is fine-tune it using the newly refined images. However, this approach may not yield the best results, as the generated content might still exhibit color and geometry inconsistencies compared to the training views.
Therefore, a constraint derived from the real data is necessary. As a result, we also incorporate the reliable pixels obtained during the refinement process into the fine-tuning procedure.

Another crucial aspect of 3DGS fine-tuning is the densification of Gaussian primitives. Increasing the density of primitives enhances the model's capability in modeling complex texture regions. In practice, we apply Gaussian primitive densification only when the to-be-updated views are within $\frac{1}{3}\times$ the original distance. This scheme enables the model to gradually increase the number of Gaussian primitives as more views are rendered and incorporated as new training data.

\vspace{-2pt}

\section{Experiments}
\label{sec:exp}

\begin{figure*}[!t]
\centering
\includegraphics[width=0.96\linewidth]{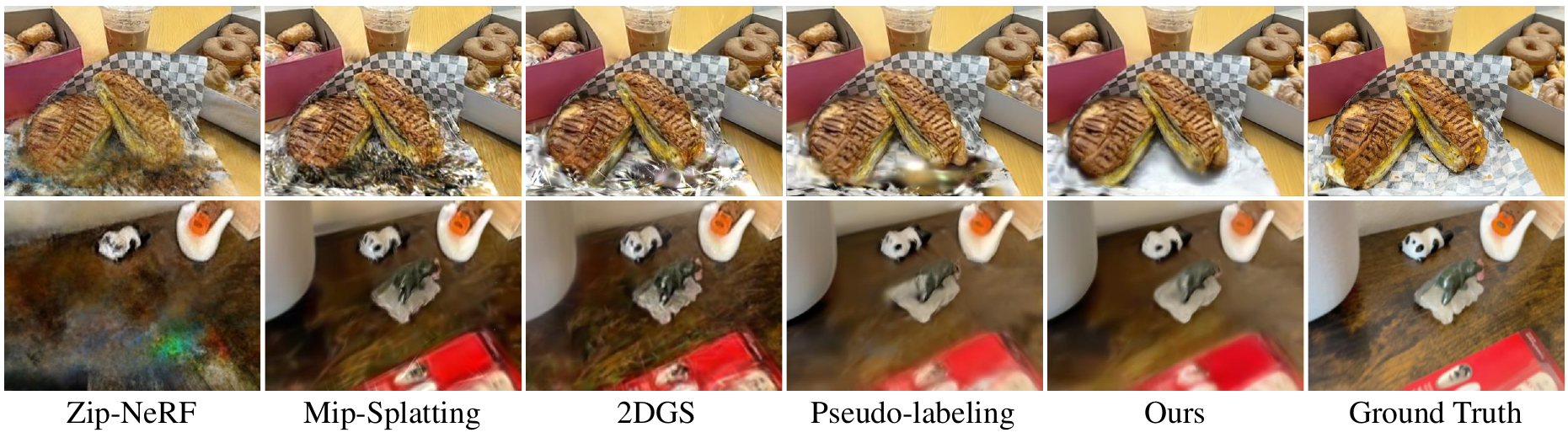} 
\vspace{-4pt}
\caption{Qualitative comparisons with representative methods on extracted LERF dataset.
}
\vspace{-2pt}
\label{fig:lerf}
\end{figure*}

In Sec.~\ref{sec:dataset}, we provide a detailed description of the scenes used, with a particular emphasis on close-up views and the evaluation metrics for this issue. In Sec.~\ref{sec:imple}, we present our implementation details.  In Sec.~\ref{sec:comparison}, we show the results of our method and other widely used methods. In Sec.~\ref{sec:ablation}, we conduct ablation studies on the  proposed methods.

\begin{table}[!t]
\vspace{3pt}
\begin{center}
{
\resizebox{1\linewidth}{!}{
\begin{tabular}{l|ccccc}
\hline
Methods & PSNR$\uparrow$ & SSIM$\uparrow$ & LPIPS$\downarrow$ & DINO$\uparrow$ & MetaIQA$\uparrow$  \\
\hline

NeRF~\cite{mildenhall2020nerf} &  $14.93$  &  $0.569$  &  $0.623$  &  $0.767$  &  $0.275$ \\
Mip-NeRF~\cite{barron2021mip} &  $16.49$  &  $0.577$  &  $0.623$  &  $0.799$  &  $0.269$ \\
Zip-NeRF~\cite{barron2023zipnerf} &  $16.54$  &  $0.583$  &  $0.565$ &  $0.803$   &  $0.309$ \\
3DGS~\cite{kerbl3Dgaussians} &  $16.22$  &  $0.558$  &  $0.560$ &  $0.814$  &  $0.353$ \\
2DGS~\cite{Huang2DGS2024} &  $16.75$  &  $0.600$  &  $0.538$ &  $0.847$  &  $0.376$ \\
GaussianPro~\cite{cheng2024gaussianpro} &  $16.09$  &  $0.548$  &  $0.563$ &  $0.813$  &  $0.308$ \\
Deblurring3DGS\cite{lee2024deblurring} &  $16.11$  &  $0.545$  &  $0.574$ &  $0.793$  &  $0.325$ \\
DNGaussian~\cite{li2024dngaussian} &  $16.28$  &  $0.599$  &  $0.605$ &  $0.769$  &  $0.310$ \\
Mip-Splatting~\cite{Yu2024MipSplatting} &  $16.69$  &  $0.591$  &  $0.542$  &  $0.826$ &  $0.346$\\
Pseudo-labeling~\cite{xia2025pseudo}   &  $17.48$  &  $0.628$  &  $0.513$  &  $0.881$ &  $0.403$ \\

\hline
Ours &  $\mathbf{17.94}$  &  $\mathbf{0.640}$  &  $\mathbf{0.502}$ &  $\mathbf{0.886}$ &  $\mathbf{0.431}$ \\

\hline
\end{tabular}
}
\caption{Quantitative comparisons on extracted LERF dataset.}
\vspace{-4pt}
\label{tab:cmp_lerf}
}
\end{center}
\end{table}

\begin{figure*}[!t]
\centering
\includegraphics[width=1\linewidth]{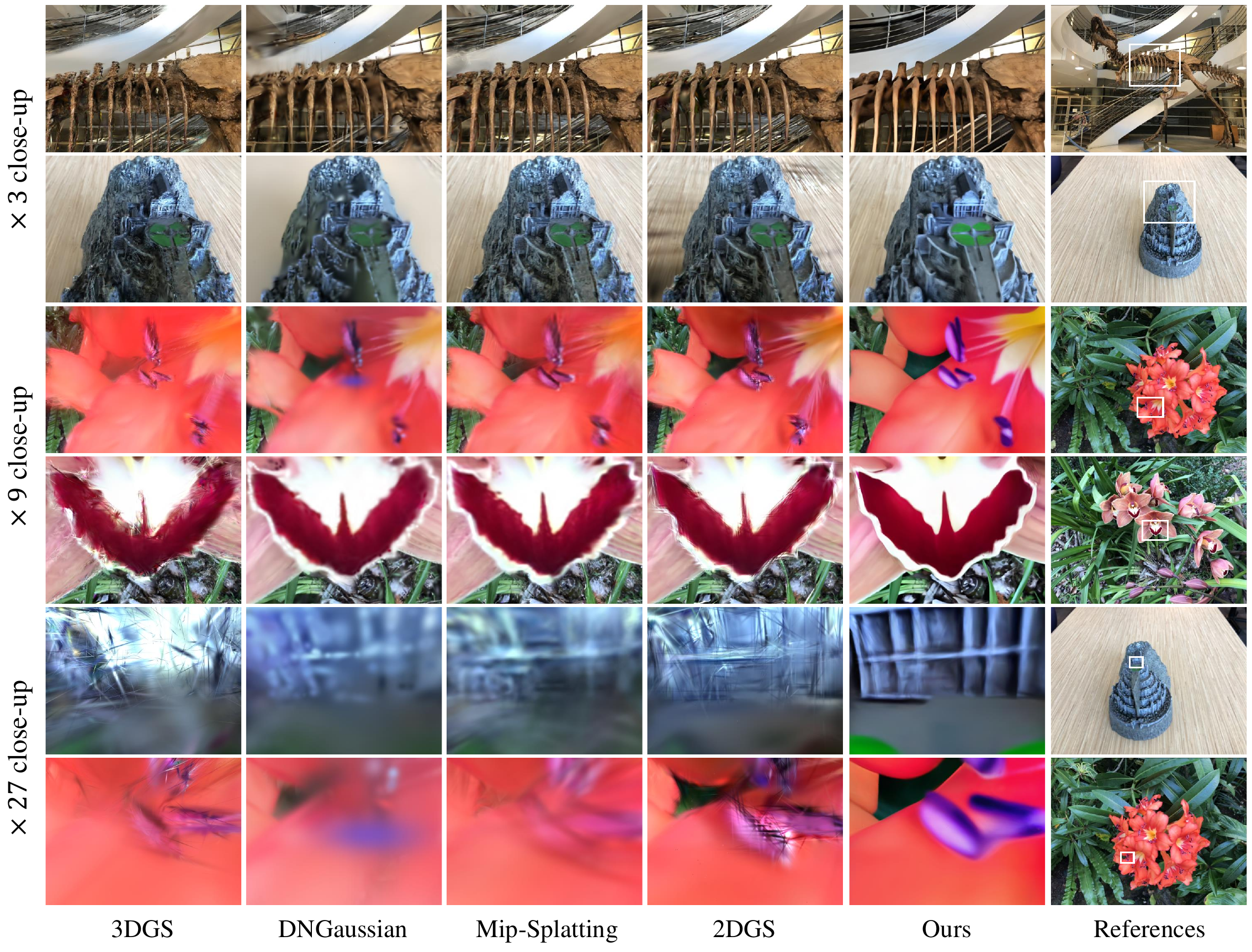} 
\vspace{-13pt}
\caption{Comparisons of progressive update on LLFF dataset with 3 close-up scales, each scale showing two rows of results. The far distance reference images are displayed on the right, with the white box outlines the approximate close-up region.
}
\label{fig:llff}
\end{figure*}

\subsection{Close-up Views Data and Metrics}
\label{sec:dataset}

\paragraph{Extracted LERF dataset.}
Existing widely used datasets do not specifically focus on the issue of close-up views. In order to have numerical results compared to ground-truth images, we chose to extract scenes specifically tailored to close-up views from the LERF dataset~\cite{kerr2023lerf}. LERF dataset is proposed for open-vocabulary tasks, and its scenes include both distant views and close-up views of objects, and we choose views from the long-distance as training views, and views from the close-up distance as testing views.
We conducted a statistic analysis of the distances between the object and the views. The testing views are mainly positioned at a distance 3$\times$ closer than that between the object and the training views. This fits well as one round in the progressive update process.
Therefore, we utilize the extracted LERF dataset for single-round update experiments, accompanied by ground-truth comparisons.
\vspace{-7pt}

\paragraph{Close-up on LLFF dataset.}
We present the close-up experiments on the most widely used novel view synthesis dataset, LLFF~\cite{mildenhall2019llff}. We selected an object-of-interest in different scenes, and conduct comparisons on three close-up scales: 3$\times$, 9$\times$, and 27$\times$. This involves using the metrics we proposed for numerical comparison.

\paragraph{Close-up views evaluation metrics.}
When ground truth images are available, we use standard metrics from novel view synthesis, including PSNR, SSIM~\cite{ssim}, LPIPS~\cite{lpips}. However, obtaining ground-truth images for most close-up view scenarios is challenging, especially for extreme close-ups that are difficult to capture in practice. Therefore, we introduce two new metrics to evaluate close-up views. 
One metric involves using the training image with the most coverage of the close-up objects as a reference to evaluate the DINO~\cite{oquab2023dinov2} score of the close-up views. It reflects the correlation between the rendered close-up images and the original images.
We also select a no-reference image quality assessment metric MateIQA~\cite{Zhu2020MetaIQA} to obtain a numerical score for the quality of the rendered close-up view images. These two metrics deliver an objective evaluation of close-up views without requiring ground-truth images.
\vspace{-2pt}

\subsection{Implementations}
\label{sec:imple}
\vspace{-2pt}

We divided our progressive update into 3 rounds, progressively moving the frontier views 3$\times$, 9$\times$, and 27$\times$ closer to the original distance to $P_{target}$ (each round moves 3$\times$ closer compared to the previous round), achieving an extreme close-up at 27$\times$ closer distances. For the view selection at each round, we use our selection method to choose 5 reference views and 8 to-be-updated views. For single-round update on extracted LERF dataset, we apply the testing views as the frontier views to run our method for one round. We choose the 30000 iterations initial optimized 2DGS~\cite{Huang2DGS2024} using training views as our baseline method. In each update round, the refined samples, along with the reliable pixel samples obtained during the refine process, are added to the existing training samples, to fine-tune the Gaussian Splatting for 5000 iterations. And those update views with a distance 3$\times$ closer are used to densify the Gaussian primitives. We conduct our experiments on  NVIDIA 4090 GPUs.

\vspace{-2pt}

\begin{table}[!t]
\vspace{5pt}
\begin{center}
{
\resizebox{0.99\linewidth}{!}{
\begin{tabular}{l|cc|cc}
\hline
   & \multicolumn{2}{c|}{Last Round~(27$\times$ closer)} & \multicolumn{2}{c}{Overall} \\
Methods   & ~~DINO$\uparrow$ & MetaIQA$\uparrow$ & ~DINO$\uparrow$ & MetaIQA$\uparrow$   \\
\hline
3DGS~\cite{kerbl3Dgaussians} &  $0.580$   &  $0.287$  &  $0.695$ &  $0.344$\\
2DGS~\cite{Huang2DGS2024} &  $0.578$  &  $0.279$  &  $0.703$ &  $0.341$\\
GaussianPro~\cite{cheng2024gaussianpro} &  $0.573$  &  $0.272$  &  $0.694$  &  $0.321$\\
Delurring3DGS~\cite{lee2024deblurring} &  $0.598$  &  $0.291$  &  $0.706$  &  $0.338$\\
DNGuassian~\cite{li2024dngaussian} &  $0.605$  &  $0.260$  &  $0.703$ &  $0.301$   \\
Mip-Splatting~\cite{Yu2024MipSplatting} &   $0.601$ &  $0.254$  &  $0.724$ &  $0.313$ \\
Pseudo-labeling~\cite{xia2025pseudo}   &  $0.586$   &  $0.265$  &  $0.718$  & $0.343$ \\

\hline
Ours w/o progressive   &  $0.587$  &  $0.322$  &  $0.728$  &  $0.387$  \\
Ours w/o densify   &  $0.602$  &  $0.324$  &  $0.721$  &  $0.391$  \\
Ours &  $\mathbf{0.616}$  &  $\mathbf{0.360}$  &  $\mathbf{0.730}$ &  $\mathbf{0.411}$ \\

\hline
\end{tabular}
}
\caption{Comparisons of progressive update on LLFF dataset. The last round is the 27$\times$ close-up update round.}
\vspace{-16pt}
\label{tab:cmp_llff}
}
\end{center}
\end{table}

\subsection{Main Results}
\label{sec:comparison}
\vspace{-2pt}

\paragraph{Single-round update on extracted LERF dataset.}
Tab~\ref{tab:cmp_lerf} shows the comparison under close-up views with widely used methods, our method shows significant improvements in all metrics compared to other methods. Specifically, our method outperforms others in traditional metrics including PSNR, SSIM, and LPIPS, highlighting its ability to produce high-quality close-up images with strong consistency. The substantial improvement on MetaIQA further demonstrate the clear enhancement in image quality achieved by our method. 
We show visualization comparisons with some representative methods in Fig.~\ref{fig:lerf}, especially in row 1, where other methods including Pseudo-labeling~\cite{xia2025pseudo} exhibited issues with artifacts. Our method effectively addresses this issue, demonstrate a notable improvement in close-up views rendering quality, presenting more complete and smoother results.
\vspace{-8pt}

\begin{figure*}[!t]
\centering
\includegraphics[width=0.83\linewidth]{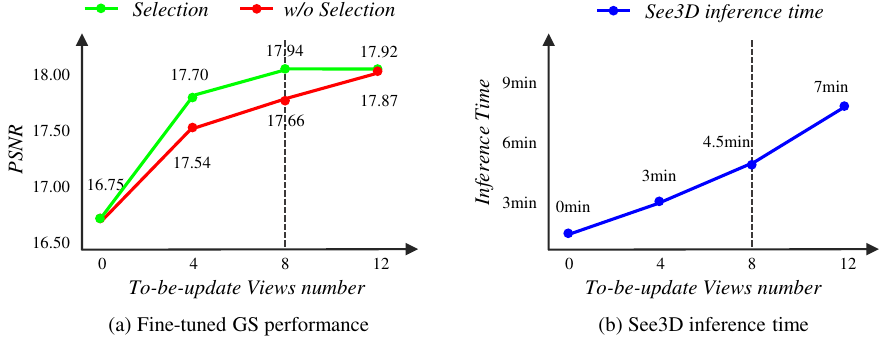} 
\vspace{-2pt}
\caption{Ablation studies of views selection. (a) shows a single-round updated Gaussian Splatting's PSNR performance on LERF dataset using same number of anchor views and different number of to-be-update views. (b) shows the corresponding See3D's inference time. 
}
\vspace{3pt}
\label{fig:selection}
\end{figure*}

\paragraph{Progressive close-up update on LLFF dataset.}
Fig.~\ref{fig:llff} shows visualization results of views in three different close-up scales. As close-up scale increases, rendering issues from other methods become more evident, whereas our approach maintains excellent rendering quality at all scales. Even at a 27$\times$ close-up distance, it still produces impressive results, as seen in the reference images, the views refer to a very small area in the original training image, yet our method effectively handles it. 
This can also be observed in Tab.~\ref{tab:cmp_llff}, for the 27$\times$ close-up scale, our method achieves a significant lead in image quality MetaIQA. In the overall evaluation, our method also demonstrates a clear and significant lead.

\subsection{Ablation Studies}
\label{sec:ablation}

\paragraph{Ablation studies of refine methods.}
We evaluate the improvement of different refine methods in Tab.~\ref{tab:ablation_}. 
It is important to note that the See3D results show sub-par performance in PSNR and SSIM, indicating the issue with consistency between the See3D results and the real data.
Directly fine-tuning with the See3D results reduces some of the inconsistencies, but there is still a significant gap from the optimal result.
Reliable pixels can lead to good fine-tuning results, since they are sparse, they cannot achieve the optimal outcome.
The combination of reliable pixels and See3D results achieve excellent fine-tuning, with the subsequent enhancement leading to the best performance. 
It should be noted that directly enhancing the reliable pixels does not improve the results; the optimal solution is to enhance the See3D results and use reliable pixels to correct the inconsistencies.
\vspace{-1mm}

\begin{table}[!t]
\vspace{5pt}
\begin{center}
{
\resizebox{1\linewidth}{!}{
\begin{tabular}{l|ccc}
\hline
    &\multicolumn{3}{c}{LERF dataset} \\ 
Methods   & ~PSNR$\uparrow$~ & ~SSIM$\uparrow$~ & ~LPIPS$\downarrow$~  \\
\hline
Baseline~\cite{Huang2DGS2024} &  $16.75$ &  $0.600$  &  $0.538$ \\
See3D render &  $16.35$  &  $0.566$  & $0.492$   \\
See3D update&  $17.22$  &  $0.614$ & $0.517$  \\
Reliable Pixels update &  $17.65$  &  $0.635$  & $0.514$ \\
Reliable Pixels + Enhance update&  $17.48$  & $0.621$  &  $0.519$  \\
Reliable Pixels + See3D update &  $17.86$  &  $0.636$ & $0.506$  \\
Reliable Pixels + See3D + Enhance update&  $\mathbf{17.94}$  & $\mathbf{0.640}$  &  $\mathbf{0.502}$ \\

\hline
\end{tabular}
}
\caption{Ablation studies of refine methods. ``See3D render'' are the results from See3D, while others use rendered view to update the model. ``Reliable Pixels'' refers to the reliable pixels obtained during the refine process, and ``Enhance'' is the image enhancement operation.
}
\vspace{-2mm}
\label{tab:ablation_}
}
\end{center}
\end{table}

\paragraph{Ablation studies of views selection.}
Fig~\ref{fig:selection} shows the effectiveness of our views selection strategy. Compared to updates without view selection, our method enables the update process to fully utilize the limited number of views, achieving better results in much less See3D inference time without views selection. This significantly improves update efficiency, ensuring the best outcomes with minimal cost.
\vspace{-1mm}

\paragraph{Ablation studies of Gaussian Splatting update.}
As shown in Tab.~\ref{tab:cmp_llff}, compared to the fine-tuning without progressive updates and without Gaussian primitives densify, our method achieves higher overall correlation and image quality. Especially in the extreme close-ups (last round), both factors play a crucial role in delivering a noticeable improvement for our approach.

\section{Conclusion} 
This work effectively addresses the long-standing challenge in existing Gaussian Splatting methods of generating reliable images from close-up views. We propose an approach that leverages existing generative models to refine close-up novel views and apply them to progressively expand the Gaussian Splatting's ``trust region''. A proposed views selection strategy ensures this process in a cost-effective manner. Additionally, we introduce a Gaussian Splatting fine-tuning procedure to ensure high-quality optimization in the close-up regions. We also define new metrics and specific scenes to better evaluate close-up views. Overall, our approach demonstrates a robust and effective solution for handling close-up views problem.

\clearpage

{
    \small
    \bibliographystyle{ieeenat_fullname}
    \bibliography{main}
}

\clearpage

\appendix

\section{Views Selection Algorithm}
We detailed the proposed views selection algorithm in Algorithm~\ref{alg:view_seletct}, which is a greedy initialization local searching algorithm, as we select the set of nodes that have highest weight as the initial solution and iteratively swap the node in this solution to find the maximum $s^{T}\cdot\mathbf{w} - s^{T}\cdot \mathbf{E}\cdot s$.  ultimately identifying $K$ optimal elements as the selected views.
We designed it as an optimization algorithm because the number of views making brute-force searching for views has a very high time complexity ($O(N^{K})$). Therefore, a highly efficient optimization algorithm with low time complexity ($O(1)$) is needed to make this process feasible. We set the optimization to run for 1,000 iterations in our paper's case, ensuring that the entire algorithm can perform real-time computation.

\begin{algorithm}[ht]
\caption{Views selection optimization}\label{alg:view_seletct}
\begin{algorithmic}[1]
\Require Weights $\mathbf{w}$; Similarity $\mathbf{E}$; Selected Number $k$
\Ensure Maximizing $s^{T}\cdot\mathbf{w} - s^{T}\cdot \mathbf{E}\cdot s$.
\State \textbf{Initialize:} Sort $\mathbf{w}$ and select top $k$ elements as $selected\_w$
\State Set a binary vector $s$ with same size of $\mathbf{w}$
\State Set the elements of $s$ with $selected\_w$’s position as 1, other as 0.
\State  Compute initial $best\_value = s^{T}\cdot\mathbf{w} - s^{T}\cdot \mathbf{E}\cdot s$
\State Set $max\_iter = 1000$
\For{$iter = 1$ to $max\_iter$}
 \State Randomly pick $w \in selected\_w$
 \State Randomly pick $new\_w \in (\mathbf{w} - selected\_w)$
 \State Swap $w$ with $new\_w$ in $selected\_w$
 \State update $s$'s elements with $selected\_w$
 \State \textbf{Compute new value:}
    \State $new\_value = s^{T}\cdot\mathbf{w} - s^{T}\cdot \mathbf{E}\cdot s$
\State \textbf{Acceptance criteria:}
\If{$new\_value > best\_value$}
\State Update $best\_value = new\_value$
\Else
\State Revert swap
\EndIf
\EndFor
\State \Return  selection vector $s$
 \end{algorithmic} 
\end{algorithm}

\section{Camera Operations}
Many of the operations in our paper such as determining the target view, are based on the changes of camera poses $P=[R|T]$ with its focal length $f$. The two main camera operations involved are changing the orientation and move closer. 

\paragraph{Orientation.} When the image coordinates $(i,j)$ in a $h\times w$ of a point-of-interest is known, and we want to adjust the camera's orientation so that this point is positioned at the center of the image, we can first change the rotation matrix $R$ of the camera into the Euler angles $(\theta_{x},\theta_{y},\theta_{z})$:
\begin{equation}
    (\theta_{x},\theta_{y},\theta_{z})=F_{R\to E}(R)
\end{equation}
Then compute the angle at which this point appears with the camera orientation:
\begin{equation}
\begin{aligned}
        \Delta \theta_{x}=\arctan(\frac{i-\frac{h}{2} }{f} ),\\
    \Delta \theta_{y}=\arctan(\frac{j-\frac{w}{2} }{f} ).   
\end{aligned}
\end{equation}
And we can change the Euler angles of the camera orientation as:
\begin{equation}
    \left\{\begin{matrix}
\theta_{x}^{'}= \theta_{x}+\Delta \theta_{x}
 \\ \theta_{y}^{'}= \theta_{y}+\Delta \theta_{y}
 \\\theta_{z}^{'}= \theta_{z}
\end{matrix}\right.
\end{equation}
Then change the Euler angles into the new rotation matrix:
\begin{equation}
    R'=F_{E\to R}(\theta_{x},\theta_{y},\theta_{z})
\end{equation}
Thus, we can obtain the new camera pose with the point-of-interest as the image's center:
\begin{equation}
    P'=[R'|T]
\end{equation}

\paragraph{Move closer.} Using a close-up factor $0 < \alpha_t < 1$ we can move the camera $P$ closer to a target position $P_{target}$:
\begin{equation}
    P''=(1-\alpha_{t})\cdot P+\alpha_{t}\cdot P_{target}
\end{equation}

\section{Discount Factor in Observability}
\begin{figure}[!t]
\centering
\includegraphics[width=1\linewidth]{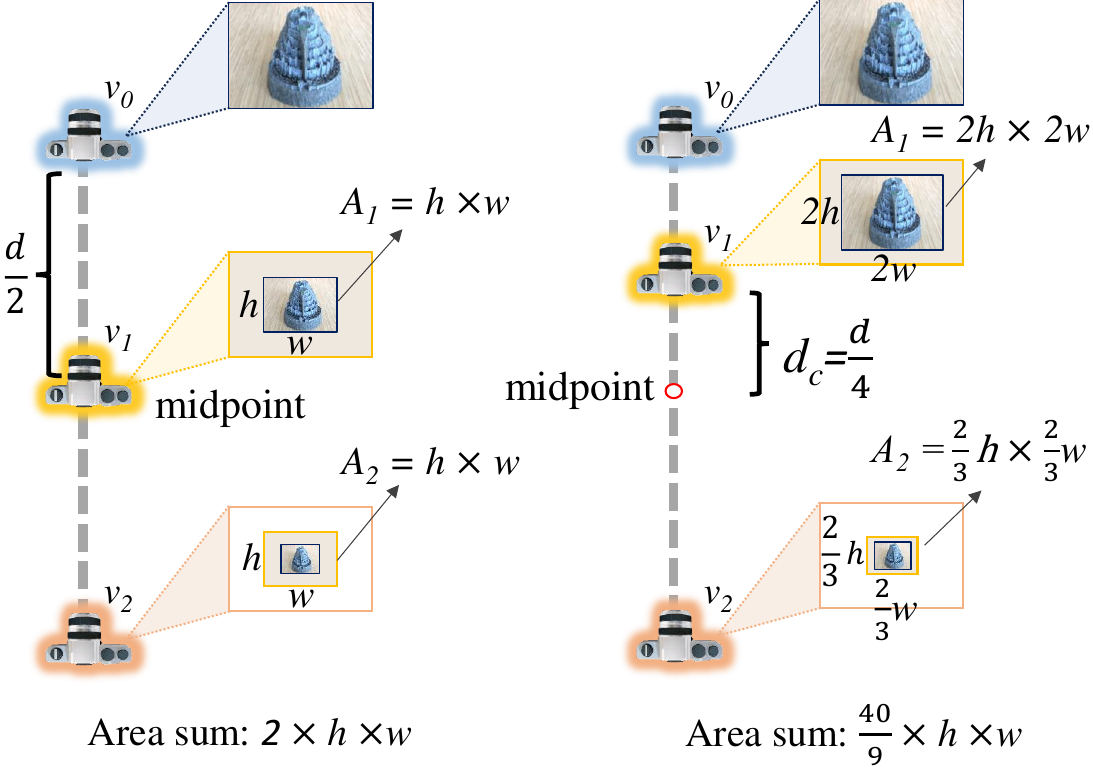} 
\caption{Distance factor in observability. Here we show two different positions of view $v_{1}$, and the sum of the area $A_{1}$ which is the observability of view $v_{0}$ under view $v_{1}$ and $A_{2}$ which is the observability of view $v_{1}$ under view $v_{2}$ reaches its minimum when $v_{1}$ is positioned in the midpoint of $v_{1}$ and $v_{2}$, and varies with the change of $d_{c}$ which is the distance between $v_{1}$ and the midpoint.}
\label{fig:distance}
\end{figure}

When we want to comprehensively evaluate the contents a view $v_{0}$ observe is visible to $v_{1}$ (which is the observability of $v_{0}$ under $v_{1}$) together with the content $v_{1}$ observe is visible to another view $v_{2}$, the distance between $v_{1}$ and both $v_{0}$, $v_{2}$, will naturally impact the statistical results. As shown in Fig.~\ref{fig:distance}, as the distance $d_c$ increases, the sum result of area $A_{1}$ and $A_{2}$ becomes larger compared to views with a smaller $d_c$. We define $a$ as the observability of $v_{0}$ under $v_{1}$ when $v_{1}$ is at the midpoint between $v_{0}$ and $v_{2}$. The area sum $A$ of $A_{1}$ and $A_{2}$ can then be defined as:
\begin{equation}
    A=((\frac{1}{1+d_{c}})^2 + (\frac{1}{1-d_{c}})^2)\cdot a
\end{equation}
which can be simplify as:
\begin{equation}
\begin{aligned}A= \frac{1}{\beta} \cdot a,~~~~~~~~\\
  \rm{where}~~\beta=\frac{(1-d_c^2)^2}{2\cdot(1+d_c^2)}
\end{aligned}
\end{equation}
Thus, if we want to eliminate the influence of distance on the statistical results in this situation, we need to multiply the observability results by a distance discount factor $\beta$.

\section{Datasets Details}
\paragraph{Extracted LERF dataset.} For our experiments on the extracted LERF dataset, we select three scenes from LERF dataset following the setting outlined in the paper. The three scenes are: ``donuts'', ``shoe\_rack'', ``teatime''. For each scenes, we run COLMAP on all the views that selected to calculate their camera parameters, and separate training and testing views when training the Gaussian Splatting. 

\paragraph{LLFF dataset.} 
For our progressive close-up update experiments on the LLFF dataset, we select four scenes which are: ``flower'', ``fortress'', ``orchids'', ``t-rex''. We apply all the views as training views to train the initial Gaussian Splatting. For the progressive close-up, we select object-of-interest in these scenes, specifically, the stamen in ``flower'', the building in ``fortress'', the petal in ``orchids'', the bone in ``t-rex''. We then use these object-of-interest to move the training views closer at three close-up scales 3$\times$, 9$\times$ and 27 $\times$, to obtain 5 target views at each close-up scale. These target views are used as testing views for each corresponding close-up scale.

\section{Visualization}
\begin{figure}[!t]
\centering
\includegraphics[width=1\linewidth]{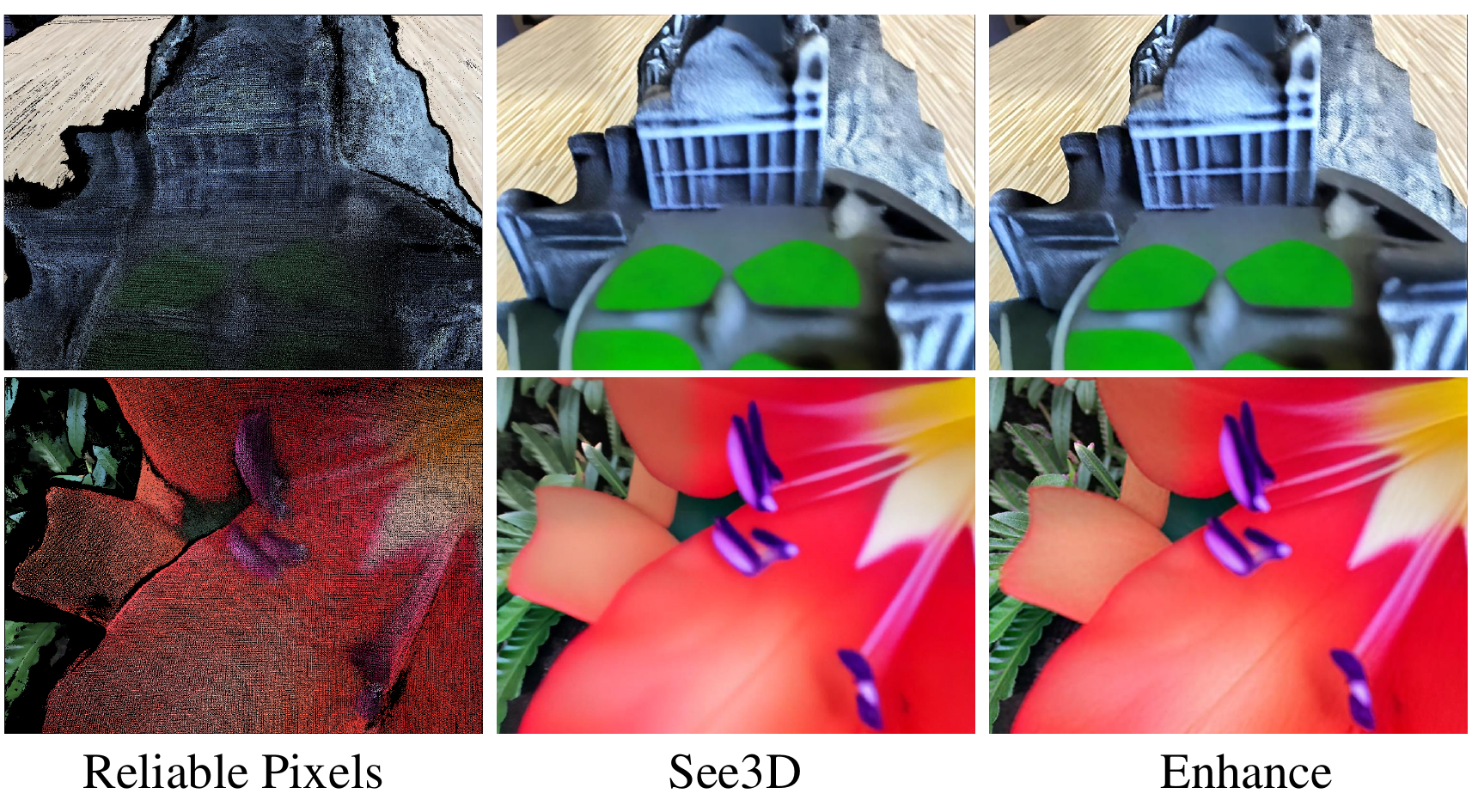} 
\caption{Refine process outcomes.
}
\label{fig:refine}
\end{figure}

\paragraph{Refine process.} 
We show two example outcomes from the refine process in Fig.~\ref{fig:refine}. 
The combination of reliable pixels and enhanced images of to-be-updated views is used to update the Gaussian Splatting.

\begin{figure}[!t]
\centering
\includegraphics[width=1\linewidth]{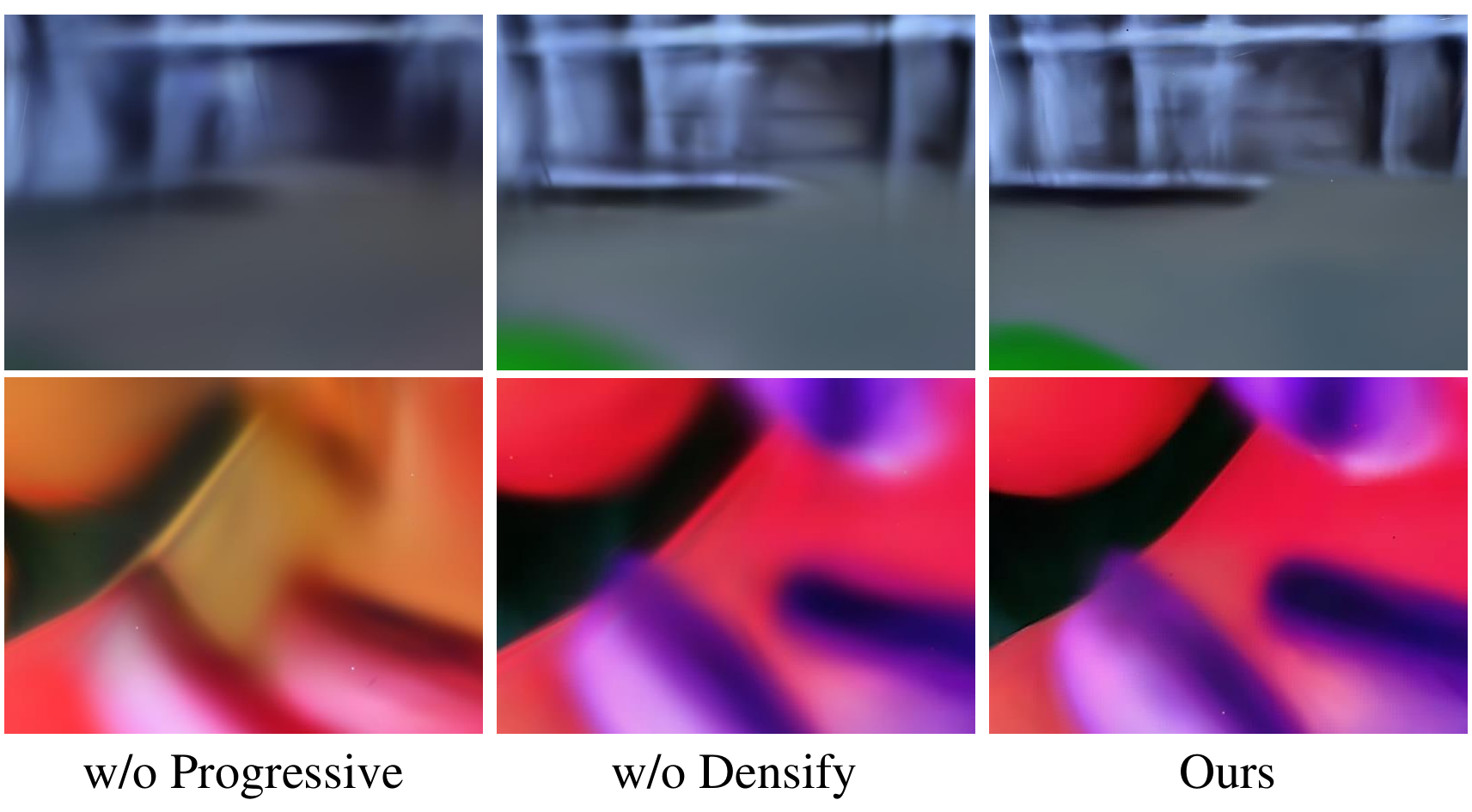} 
\caption{Visualization of Gaussian Splatting update.
}
\label{fig:ablation}
\end{figure}

\paragraph{Ablation studies of Gaussian Splatting update.}
We show the visualization result of Gaussian Splatting update ablation studies in Fig.~\ref{fig:ablation}. And it can observed that the update without progressive strategy fails to refine the close-up views, this demonstrates that each update round has its limitations, it is not feasible to achieve an extreme-scale close-up in a single round, and the progressive strategy ensures a robust updating process. Compared to updates without densify Gaussian primitives, our results exhibit greater clarity, highlighting the essential role of densifying the insufficient primitives under close-up views.

\end{document}